\documentclass[sigconf]{acmart}

\usepackage{booktabs} 

\setcopyright{rightsretained}

\acmConference[KDD '18]{KDD Workshop on Machine Learning for Medicine and Healthcare}{August 2018}{London, UK}
\acmYear{2018}
\copyrightyear{2018}

\begin{document}
\title{Measuring the quality of Synthetic data for use in competitions}
\subtitle{Extended Abstract}

\author{James Jordon}
\affiliation{%
  \institution{University of Oxford}
}
\email{james.jordon@wolfson.ox.ac.uk}

\author{Jinsung Yoon}
\affiliation{%
  \institution{University of California, Los Angeles}
}
\email{jsyoon0823@gmail.com}

\author{Mihaela van der Schaar}
\affiliation{%
  \institution{University of Oxford}
}
\email{mihaela.vanderschaar@eng.ox.ac.uk}


\begin{abstract}
	Machine learning has the potential to assist many communities in using the large datasets that are becoming more and more available. Unfortunately, much of that potential is not being realized because it would require sharing data in a way that compromises privacy. In order to overcome this hurdle, several methods have been proposed that generate {\em synthetic} data while preserving the privacy of the real data. In this paper we consider a key characteristic that synthetic data should have in order to be useful for machine learning researchers - the relative performance of two algorithms (trained and tested) on the synthetic dataset should be the same as their relative performance (when trained and tested) on the original dataset.
\end{abstract}

%
%


\maketitle
\section{Introduction}
The availability of large datasets presents enormous opportunities for collaboration between the data-holders and the machine learning community. However, many of these large datasets include sensitive information that prevents data-holders from sharing the data. In order to circumvent the problem of sharing sensitive data, several methods for generating synthetic data have been proposed \cite{ml4h_current} \cite{tstr} \cite{dpgan}.

Measuring the quality of synthetic data generation methods is difficult, however. Often the synthetic data is compared on a feature-by-feature basis with the real data to check that the 1-dimensional distributions are similar \cite{ml4h_current}. This, however, provides no information about the correctness of the joint distribution between features. Another method involves training a machine learning model on the synthetic data and then testing on the real data. If the trained model performs well, then this is indicative of the synthetic data being similar to the real data. This is an important metric when the synthetic data will be used to train models to be deployed on real data, but is not the most important metric in the setting we consider.

The performance metric we propose in this paper is grounded in a particular application of synthetic data generation. The problem here is that the data-holder wishes to find the best machine learning method to use for their task, but they do not wish to share their dataset with the many machine learning researchers that may be able to provide suitable methods. Instead, they wish to generate a synthetic dataset that they can give out to each of the machine learning researchers to train, validate and develop algorithms on. In this situation it is important than when the researchers compare two algorithms on the synthetic data, the results of the comparison are similar to those they would have obtained had they compared them on the real data. This allows the researchers to develop algorithms in a meaningful way, so that the final algorithm they select to give to the data-holder is indeed the same as the one they would have chosen had they performed the research and development on the real data itself.

It may seem that, given the preceding discussion, the only thing that is important is that the best performing algorithm on the synthetic data is also the best performing algorithm on the real data. This would be the case if the process by which an ML researcher decides on an algorithm is to test all possible algorithms simultaneously and select the best. In practice this is not the case - a researcher will develop an algorithm over time by comparing a small set of algorithms, selecting the best, and then comparing the best within another small set of algorithms. Note that this comparison may be of different variants of the "same" algorithm with hyper-parameters adjusted (or equivalently the spaces within which hyper-parameters are searched for). The researcher will then select one of the two variants based on performance on some held-out testing set and then continue to make adjustments to the new algorithm. It is therefore important that at each stage of this process, the best algorithm is selected. This means that comparisons between {\em any two} algorithms on the synthetic data should be similar to comparisons of the same two algorithms on the real data.

\section{Formalism}
\subsection{Notation}
Consider a dataset, $\mathcal{D} = \{\mathbf{u}_i\}_{i = 1}^n$ consisting of $n$ i.i.d. samples of some random variable $\mathbf{U}$ that takes values in a space $\mathcal{U}$. We consider a task, $\mathcal{T}$, that is to be performed on samples of $\mathbf{U}$. We split $\mathcal{D}$ into a training set, $\mathcal{D}_1$, and testing set, $\mathcal{D}_2$.

Let $\mathcal{A}_1, ..., \mathcal{A}_k$ be $k$ machine learning methods that take as input a (training) dataset, $\mathcal{D}_1$, and output a model $\mathcal{A}_i(\mathcal{D}_1)$ that can perform task $\mathcal{T}$ on samples of $\mathbf{U}$. Let $m_\mathcal{T}$ denote a performance metric for task $\mathcal{T}$ that takes as input a (trained) model, $\mathcal{M}$, and a (testing) dataset, $\mathcal{D}_2$, and outputs a value in $\mathbb{R}$ (i.e. $m_\mathcal{T}(\mathcal{M}, \mathcal{D}_2) \in \mathbb{R}$).

Denote by $G$ a synthetic data generation method and let $\mathcal{D}^G$ denote the (synthetic) dataset generated by it. As above we will split each synthetic dataset into training, $\mathcal{D}^G_1$, and testing, $\mathcal{D}^G_2$ sets.

\subsection{Synthetic Ranking Agreement (SRA)}
The property we are interested in can be formalized in the following way:
\begin{multline}
m_\mathcal{T}(\mathcal{A}_i(\mathcal{D}^G_1), \mathcal{D}^G_2) < m_\mathcal{T}(\mathcal{A}_j(\mathcal{D}^G_1), \mathcal{D}^G_2) \\ \implies m_\mathcal{T}(\mathcal{A}_i(\mathcal{D}_1), \mathcal{D}_2) < m_\mathcal{T}(\mathcal{A}_j(\mathcal{D}_1), \mathcal{D}_2)
\end{multline}
for all $i, j \in \{1, ..., k\}, i \neq j$.

To this end we define the Synthetic Ranking Agreement (SRA) of a synthetic data generation method, $G$.
\begin{definition}(SRA)
	Given $k$ algorithms $\mathcal{A}_1, ..., \mathcal{A}_k$ as defined above, a task $\mathcal{T}$, a performance metric $m_\mathcal{T}$, a dataset $\mathcal{D}$ and a synthetic dataset $\mathcal{D}^G$ generated by $G$, we define 
	\begin{eqnarray}
	R_i &=& m_\mathcal{T}(\mathcal{A}_i(\mathcal{D}_1), \mathcal{D}_2) \\
	S_i &=& m_\mathcal{T}(\mathcal{A}_i(\mathcal{D}^G_1), \mathcal{D}^G_2)
	\end{eqnarray}
	for each $i = 1, ..., k$ so that $R_i$ represents the performance of algorithm $i$ on the real data and $S_i$ on the synthetic data.
	The Synthetic Ranking Agreement of $G$ is then defined as
	\begin{equation}
	\textbf{SRA}(G) = \frac{1}{k(k-1)} \sum_{i = 1}^k \sum_{j \neq i} \mathbb{I} \Big( (R_i - R_j) \times (S_i - S_j) > 0 \Big).
	\end{equation}
\end{definition}
The SRA can be thought of as the (empirical) probability of a comparison on the synthetic data being "correct" (i.e. the same as the comparison would be on the real data).

Of course, its value is dependent on the set of algorithms on which the ranking is checked. Including more algorithms will generally improve the metric, however, it is naturally impossible to be exhaustive, not only because the number of existing algorithms for any given task is large and hence would be time consuming to train but also because new methods are constantly being developed. Care should also be taken in balancing the classes of methods compared - as noted above, "different" algorithms could merely be variations of the same method, such as comparing 2-layer MLP with 3-layer MLP. However, inclusion of multiple variations of a single method can artificially bias the metric - data generation methods that are particularly invariant with respect to that method will have a higher SRA than those that aren't simply because the number of incorrect orderings will be larger (despite the fact that all these incorrect orderings are coming from the same "method"). In this paper, we use 12 different prediction algorithms which we believe represent many of the common classes of prediction algorithm: Logistic Regression, Random Forests \cite{randomforest}, Gaussian Naive Bayes \cite{naivebayes}, Bernoulli Naive Bayes  \cite{naivebayes}, Linear SVM \cite{svm}, Decision Tree \cite{decisiontree}, LDA \cite{lda}, AdaBoost \cite{adaboost}, Bagging \cite{bagging}, GBM \cite{gbm}, Multi-layer Perceptron, XgBoost \cite{xgboost}.

\subsection{Privacy and SRA}
A particularly interesting property of SRA is that it does not necessarily require the synthetic data to be distributed the same as the real data to be high. This has particularly nice properties when we consider the implications this has for privacy where training synthetic data generation models to be too similar to the real data can lead to privacy concerns.

Suppose that we have a dataset consisting of $n$ feature-label pairs $(\mathbf{x}, y)$ in which the label $y$ is considered sensitive, but the features $\mathbf{x}$ are not (for example the label $y$ may represent a diagnosis). In order to make the label private, we add noise to the label. For simplicity, assume the label is binary and that the noise is that we "flip" the label with probability $p \in [0, 0.5)$. (We note that if we flip with $p = 0.5$ then the resulting data is completely random - the labels are not at all correlated with the features.)

Intuitively, the privacy of the label increases with the noise $p$ (this can also be stated and proven more formally using a notion of privacy referred to as {\em differential privacy} \cite{dpbook}). We now consider the implications of measuring the quality of such data in terms of our proposed metric and another popular metric: Training on Synthetic data and Testing on Real data (TSTR) \cite{tstr}. This latter metric is formally defined as 
\begin{equation}
\textbf{TSTR}(G) = \frac{1}{k} \sum_{i=1}^k m_\mathcal{T}(\mathcal{A}_i(\mathcal{D}^G_1), \mathcal{D}_2)
\end{equation}

\begin{figure}[b!]
	\centering
	\includegraphics[width=0.48\textwidth]{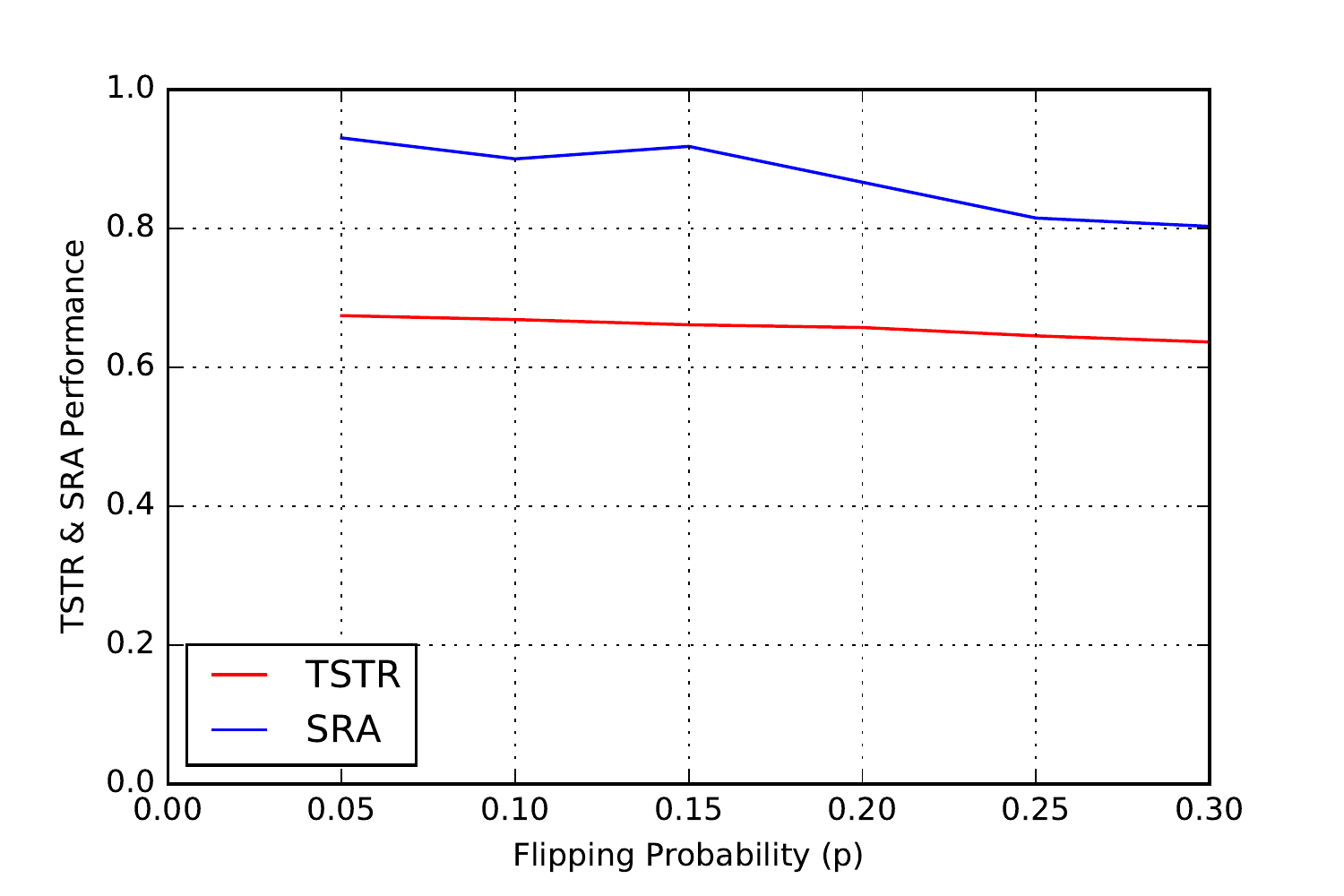}
	\caption{Comparison of SRA with TSTR}
	\label{fig:p}
\end{figure}

We compare the two metrics across the range of $p$ using the 12 predictive models given above on the MAGGIC dataset \cite{maggic}. We use AUROC as the performance metric. The results can be seen in Fig. \ref{fig:p} for $p \in [0, 0.3]$.

We note that prediction of the label by training on the noisy data is likely to be harder than from training on clean data. The quality of synthetic data as reported by TSTR is therefore going to decrease as $p$ increases. On the other hand, the relative performance of different algorithms on the noisy data does not necessarily change - the performance of {\em all} methods will be lower due to the noise. The SRA as $p$ increases is therefore not strictly decreasing as $p$ increases, as can be seen in Fig. \ref{fig:p} (from $p = 0.10$ to $p = 0.15$). This would suggest that the data generation method selected by SRA (which is most useful in the competition setting) could have stronger privacy guarantees (i.e. we can add more noise to the data) than the method selected by TSTR.

\begin{acks}
The research presented in this paper was supported by the Office of Naval Research (ONR) and the NSF (Grant number: ECCS1462245, ECCS1533983, and ECCS1407712).
\end{acks}

\subsection*{Code}
The source code is publicly available at \url{http://github.com/jsyoon0823/SRA_TSTR}.


\end{document}